\documentclass[conference]{IEEEtran}
\usepackage{ifpdf}
\usepackage{cite}
\usepackage[numbers]{natbib}
\ifCLASSINFOpdf
\else
\fi
\usepackage{array}
\usepackage{mdwmath}
\usepackage{mdwtab}
\usepackage{url}
\usepackage{graphicx}
\usepackage{times}
\usepackage{soul}
\usepackage{url}
\usepackage[hidelinks]{hyperref}
\usepackage[utf8]{inputenc}
\usepackage[small]{caption}
\usepackage{amsmath}
\usepackage{amsthm}
\usepackage{booktabs}
\usepackage{amsfonts}
\usepackage{pdfpages}
\usepackage{algorithm}
\usepackage{algorithmicx}
\usepackage{algpseudocode}

\usepackage[switch]{lineno}
\usepackage{multirow}
\usepackage{subfigure}
\begin{document}
\title{Unlearnable Graph: Protecting Graphs from Unauthorized Exploitation}

\author{\IEEEauthorblockN{Yixin Liu\\ and Lichao Sun}
\IEEEauthorblockA{Leigh University\\
\{yila22,lis221\}@lehigh.edu}
\and
\IEEEauthorblockN{Chenrui Fan\\ and Pan Zhou}
\IEEEauthorblockA{Huazhong University of Science and Technology\\
\{chenrui\_fan,panzhou\}@hust.edu.cn}
}

\author{\IEEEauthorblockN{Yixin Liu\IEEEauthorrefmark{1},
Chenrui Fan\IEEEauthorrefmark{2},
Pan Zhou\IEEEauthorrefmark{2} and
Lichao Sun\IEEEauthorrefmark{1}}
\IEEEauthorblockA{\IEEEauthorrefmark{1}
Lehigh University, Bethlehem, PA, USA}
\IEEEauthorblockA{\IEEEauthorrefmark{2}
Huazhong University of Science and Technology, Wuhan, Hubei, China\\
\{yila22, lis221\}@lehigh.edu, \{chenrui\_fan, panzhou\}@hust.edu.cn}
}

\maketitle

\begin{abstract}
%\boldmath
    While the use of graph-structured data in various fields is becoming increasingly popular, it also raises concerns about the potential unauthorized exploitation of personal data for training commercial graph neural network (GNN) models, which can compromise privacy.
    To address this issue, we propose a novel method for generating unlearnable graph examples. By injecting delusive but imperceptible noise into graphs using our \underline{\textbf{E}}rror-\underline{\textbf{Min}}imizing \underline{\textbf{S}}tructural Poisoning (EMinS) module, we are able to make the graphs unexploitable. Notably, by modifying only $5\%$ at most of the potential edges in the graph data, our method successfully decreases the accuracy from ${77.33\%}$ to ${42.47\%}$ on the COLLAB dataset. 
\end{abstract}


% \fi
%%\IEEEpeerreviewmaketitle

\section{Introduction}
The abundance of data has led to the successful implementation of deep learning, which allows the integration of artificial intelligence (AI) into various domains \cite{zhou2023comprehensive}. However, With the increasing availability of publicly accessible data, concerns have risen about the unauthorized exploitation of data. Many commercial AI models are trained using personal data that is unknowingly collected from the internet, raising questions about the potential misuse of this data for commercial or even illegal gain and also posing a significant threat to individuals' privacy, security, and copyright. 

% For example, users indict unauthorized portrait usage by a technology company for training commercial AI models of facial recognition \cite{hill2020secretive}. GitHub Copilot, which trains on trillions of open-source code without user consent, has been similarly sued \cite{sun2022coprotector}. Moreover, recently, three well-known artists are jointly suing stable diffusion over that its training materials may contain millions (or billions) of copies of copyrighted images that were obtained without the artists' knowledge or consent \cite{lang_2023}.

The threat of unauthorized data exploitation has made it imperative to develop defensive approaches. Recent studies have been focusing on developing \textit{Unlearnable Example} \cite{huang2021unlearnable,fowl2021preventing}. These methods aim to make the original data \textit{unlearnable} by adding imperceivable but delusive perturbations to data samples, resulting in deep learning models trained on the perturbed dataset having extremely low prediction accuracy. 
%While these methods share a similar objective with traditional data poisoning attacks, unlearnable examples also require that the perturbations are imperceivable, making it difficult for humans to detect the differences. As a result, the trained models are tricked into relying on these misleading and brittle perturbations, leading to poor performance in clean testing datasets and thus securing the data from unauthorized use.

Previous studies on unlearnable examples have primarily focused on the vision domain. However, as the use of graph data structures becomes more prevalent, particularly in regard to privacy and security, it is important to explore the potential vulnerability of unauthorized graph exploitation. As far as we know, unlearnable graphs, i.e., unlearnable examples on graph data, have not been explored yet. In this paper, we aim to answer the question of how to make structured graph data unlearnable by a wide range of GNN models. 

%This is a non-trivial task due to the discrete property of the graph data, which makes traditional optimization solutions from the vision domain inapplicable. In addition, compared to the vision domain, crafting unperceivable and delusive poison in structured graph data is more challenging due to the following two aspects: (1) it is difficult to maintain invisibility when only modifying inconspicuous edges or nodes \cite{sun2018adversarial}; (2) limited manipulation combinations in structural space also makes achieving delusive effects harder \cite{khalil2017learning}. 

To tackle these issues, we propose the \textit{Adaptive GradArgMin} method to craft error-minimizing structural perturbation based on the gradient information. The \textit{Adaptive GradArgMin} selects a set of edges that cause the maximum gradient change and conducts the flipping operation in the adjacent matrix. To achieve a good balance between invisibility and effectiveness under limited manipulation budgets, we design an adaptive constraint strategy by considering both vertex-based and edge-based information. The perturbed graph maintains invisible compared to the original graph under visual inspections, which ensures the utility of the data for other purposes while making them unexploitable by ML models.

\begin{figure}[t]
    \centering
    \includegraphics[width=.95\linewidth]{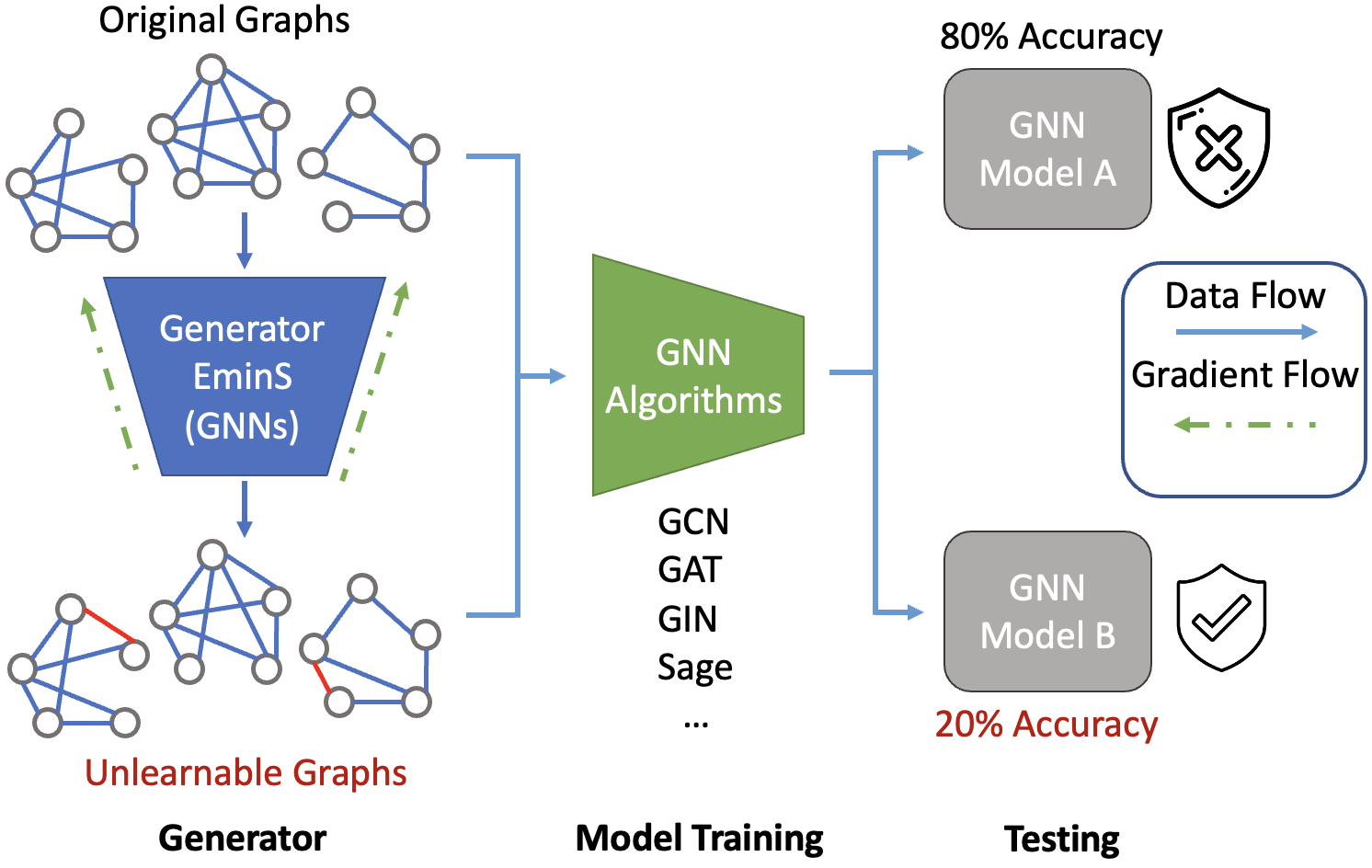}
    \caption{An illustration of motivation of Unlearnable Graph. Existing vision-based solutions fail to inject delusive patterns into more challenging data structure-graph due to their discrete property. In this paper, we propose an Error-Minimizing Structural Poisoning to achieve efficient and effective data protection for graphs.}
    \label{fig:motivation}
\end{figure}

\section{Assumptions and Problem Formulation}

\noindent\textbf{Assumptions on Defender’s Capability.}
We assume that the data owner has full access to a portion of graph data used to train a model by unauthorized data exploiters. However, the defender could not interfere with the model selection and the training procedure of the unauthorized users. 

\noindent\textbf{Objective.}
Given a clean graph training dataset $\mathcal{G}_c= \{G_i,y_i\}_{i=1}^{N}$, our goal is to craft an unlearnable version of the training dataset $\mathcal{G}_u= \{\hat{G_i},y_i\}_{i=1}^{N}$ such that the models trained on the $\mathcal{G}_u$ have poor performance on the clean testing set $\mathcal{G}_t$. The task can be formulated into a bi-level optimization as follows:
\begin{equation}
    \begin{gathered}
\max_{\delta _i\preceq \mathrm{c}} \underset{\left( \mathrm{G}_i,y_i \right) \sim \mathcal{G} _{\mathrm{t}}}{\mathbb{E}}\left[ \mathcal{L} \left( f_{\theta ^*}(\mathrm{G}_i),y \right) \right] ,\\
\,\,\mathrm{s}.\mathrm{t}.\theta ^*=\underset{\theta}{\mathrm{arg}\min}\sum_{\left( \mathrm{G}_i,y_i \right) \in \mathcal{G} _{\mathrm{u}}}{\left[ \mathcal{L} \left( f_{\theta}\left( \mathrm{G}_i\oplus \delta _i \right) ,y_i \right) \right]}.
\end{gathered}
\label{eq:bi-opt}
\end{equation}
 where $\oplus$ denotes the application of perturbations of node features or topology structure on the original graph $G_i$, and $\preceq$ represents the budget constraints relationship.
\section{Proposed Methodology}
In this section, we propose Error-Minimizing Structural noise, which is effective and imperceptible against unauthorized exploitation.

\noindent\textbf{The Min-min Optimization.}
To tackle the intractable bi-level problem in Eq. \ref{eq:bi-opt}, an approximated \textit{min-min} optimization process is proposed \cite{huang2021unlearnable} to first learn a noise generator and leverage it to conduct noise generation.
The major motivation is to iteratively craft noise that can trick the models trained on the poisoned data. The problem is also a bi-level optimization problem, with two levels of minimization. The inner level is a constrained optimization problem that finds the noise that is bounded by certain constraints and minimizes the model's classification loss. The outer level is another minimization problem that finds the parameters that also minimize the model's classification loss.

\noindent\textbf{Crafting Delusive Edges with Adaptive GradArgMin.}
The core of our method is to take gradients with respect to the adjacent matrix $A$ to obtain the gradient for any edge in the potential edge space ($\mathcal{V}\times\mathcal{V}$) regardless of its existence. 
% Specifically, for the element $\alpha_{u, v} \in A$, the gradient is computed with $\frac{\partial \mathcal{L}}{\partial \alpha_{u, v}}=\sum_{k=1}^K \frac{\partial \mathcal{L}^{\top}}{\mu_k} \cdot \frac{\partial \mu_k}{\partial \alpha_{u, v}}$.
For one selected edge $(u,v)$, we conduct the discrete version of the gradient descend 
% \ie, $\alpha_{u, v} \leftarrow \alpha_{u, v}-\eta \frac{\partial \mathcal{L}}{\partial \alpha_{u, v}}$, 
by deleting existing edges with a positive gradient or adding non-exist edges with a negative gradient. 
% We term such operation as flipping since the process is conducted by simply changing the state of edge with ${\alpha _{u,v}}^{\prime}=\mathbb{I} \left( \alpha _{u,v}=0 \right)$. 
% To be more specific, 
Given the modification constraint of edges $c$ and the element $\alpha_{u_t,v_t}\in A$, we obtain a set of edges via a greedy selection:
\begin{equation}
    \left\{ u_t,v_t \right\} _{t=1}^{\mathrm{c}}=\underset{\left\{ u_t,v_t \right\} _{t=1}^{\mathrm{c}}}{\mathrm{arg}\max}\sum_{t=1}^{\mathrm{c}}{|\frac{\partial \mathcal{L}}{\partial \alpha _{u_t,v_t}}|}.
    \label{eq:top-c}
\end{equation}
After that, modifications are performed by sequentially modifying these edges in the way that is most likely to reduce the loss function. Note that we stop the modification process until we find all the gradients for existing edges are negative, or the ones for non-exist edges are positive, in which case no perturbation for decreasing the objective is possible. 
% \begin{equation}
% \hat{G}^{t+1}=\left\{ \begin{array}{l}
% 	\left( \hat{V}_t,\hat{E}_t\backslash \left( u_t,v_t \right) \right) :\frac{\partial \mathcal{L}}{\partial \alpha _{u_t,v_t}}>0\\
% 	\left( \hat{V}_t,\hat{E}_t\cup \left\{ \left( u_t,v_t \right) \right\} \right) :\frac{\partial \mathcal{L}}{\partial \alpha _{u_t,v_t}}<0\\
% \end{array} \right. .
% \label{eq:graph-upd}
% \end{equation}

\noindent\textbf{Adaptive Constrains.}
To ensure that we modify \textit{sufficient and essential} edges in each graph for creating \textit{delusive and invisible} patterns, we devise a mixed type of constraint based on \textit{vertex-based} and \textit{edge-based} information.  We consider two types of perturbation ratios, $r_V$ and $r_E$, which refer to the entire edge space ($V\times V$) or existing edges ($E$), respectively. The number of edges to be modified is constrained by both coefficients.
% Compared to setting a hard threshold of edge number regardless of the property of each graph, 
Our constraints are effective and flexible, as they allocate more budget to larger graphs, ensuring the overall imperceptibility and effectiveness of our method.
\section{Experiments}
\begin{figure}[h]
\centering
    \begin{minipage}{0.23\textwidth}
    \centering
    \includegraphics[width=\textwidth]{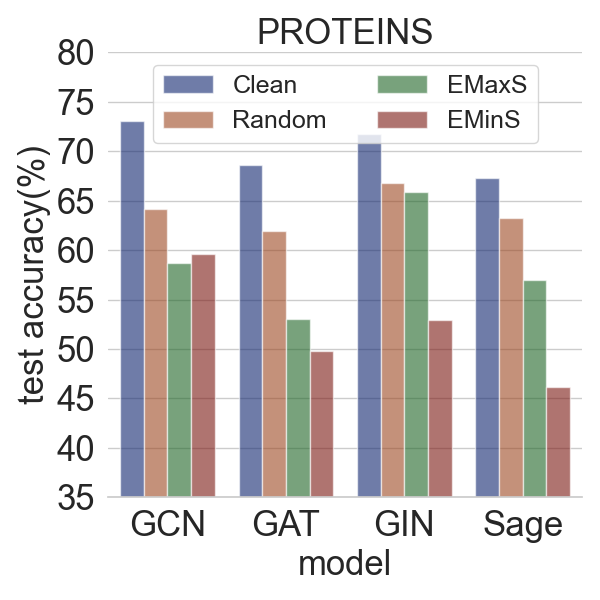}
    \end{minipage}
    \begin{minipage}{0.23\textwidth}
    \centering
    \includegraphics[width=\textwidth]{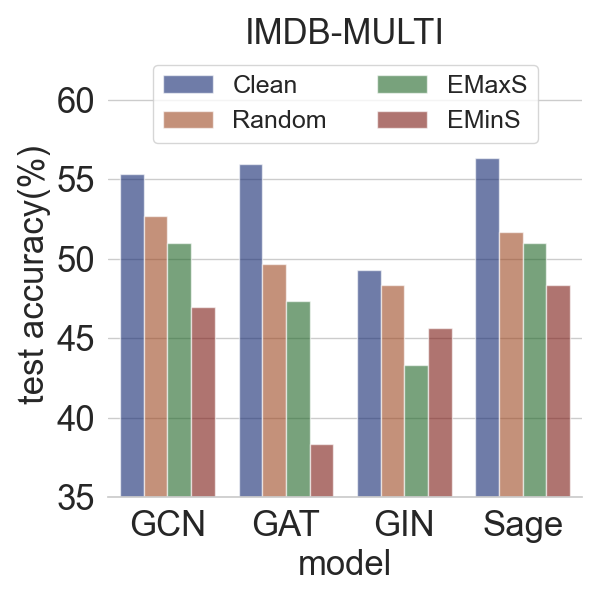}
    \end{minipage}
    \caption{A comparison among different methods on PROTEINS and IMDB-MULTI datasets.}
    \label{fig:result}
\end{figure}

In order to evaluate the effectiveness of our proposed method, we conducted experiments on six benchmark graph classification datasets(MUTAG, ENZYMES, PROTEINS, IMDB-B, IMDB-M, and COLLAB) across four common GNN architectures(GCN, GAT, GIN, and GraphSage), and make comparisons between random and error-maximizing noise. The results on the PROTEINS and IMDB-M datasets are reported in Figure \ref{fig:result}.

 Despite the good performance of EMinS noise in degrading the models' test accuracy, we visualize the graphs before and after perturbation in Table \ref{fig:visual} to demonstrate the imperceptibility of our noise. From the visualizations, we can observe that for the majority of graphs, there are few visual discrepancies between the original and modified ones. 

 \begin{table}[h]
    \centering
    \resizebox{1.01\linewidth}{!}{
    \begin{tabular}{c|cc}
        % \toprule
        &IMDB-M&ENZYMES\\
        \midrule
        \rotatebox{90}{Clean}&  \includegraphics[width=0.6\linewidth]{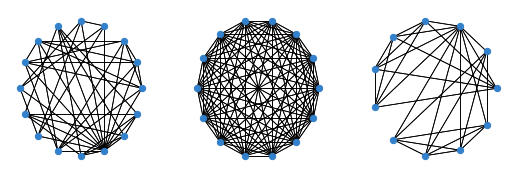}&\includegraphics[width=0.6\linewidth]{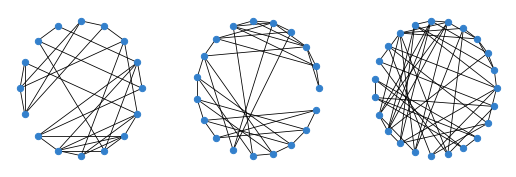}\\
        \rotatebox{90}{Perturbed} &  \includegraphics[width=0.6\linewidth]{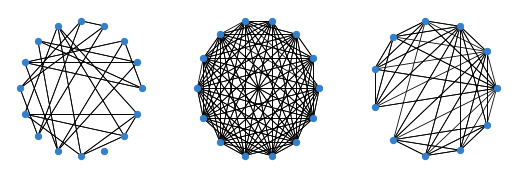}&\includegraphics[width=0.6\linewidth]{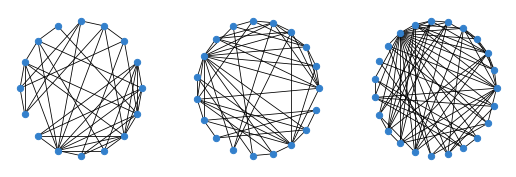}
    % \bottomrule
    \end{tabular}
    }
    \caption{Graph visualizations on IMDB-MULTI and ENZYMES datasets (the first row are clean graphs, and the second row are the graphs we generated with EMinS).}
    \label{fig:visual}
\end{table}

\section{Conclusion}
In this paper, we are the first group that proposes a novel method for minimizing errors in structural poisoning for generating unlearnable graphs. Our method explores invisible noise to prevent GNN models from exploiting graph data freely. We verify our method by conducting experiments on six benchmark graph datasets, and the extensive experimental results show that our method can be applied effectively to various GNN architectures. This study represents an important first step in safeguarding personal graph data from being exploited by GNN models.





%\bibliographystyle{IEEEtranS}


% \end{thebibliography}
\small
\bibliographystyle{named}
\bibliography{NDSS23}
\clearpage
%\includepdf[pages=-]{NDSS2023.pdf}

\end{document}